\documentclass[conference]{IEEEtran}
\IEEEoverridecommandlockouts
\usepackage{cite}
\usepackage{amsmath,amssymb,amsfonts}
\usepackage{algorithmic}
\usepackage{graphicx}
\usepackage{textcomp}
\usepackage{xcolor}
\usepackage{subcaption}

\def\BibTeX{{\rm B\kern-.05em{\sc i\kern-.025em b}\kern-.08em
    T\kern-.1667em\lower.7ex\hbox{E}\kern-.125emX}}
\begin{document}

\title{Federated Learning for Deforestation Detection: A Distributed Approach with Satellite Imagery}

\author{
\makebox[\textwidth][c]{%
\begin{minipage}[t]{0.3\textwidth}
\centering
Yuvraj Dutta\\
\textit{Electronics and Communication} \\
\textit{Engineering} \\
\textit{NIT Rourkela, India} \\
dattayuv.nitrkl@gmail.com
\end{minipage}
\hfill
\begin{minipage}[t]{0.3\textwidth}
\centering
Aaditya Sikder\\
\textit{Electronics and Communication} \\
\textit{Engineering} \\
\textit{NIT Rourkela, India} \\
aadityasikder@gmail.com
\end{minipage}
\hfill
\begin{minipage}[t]{0.3\textwidth}
\centering
Basabdatta Palit\\
\textit{IEEE Member} \\
\textit{Electronics and Communication} \\
\textit{Engineering} \\
\textit{NIT Rourkela, India} \\
palitb@nitrkl.ac.in
\end{minipage}
}
}

\maketitle

\footnotetext[1]{\label{foot:flower}https://flower.ai/docs/framework/index.html}
\footnotetext[2]{\label{foot:ray}https://www.ray.io}

\begin{abstract}
Accurate identification of deforestation from satellite images is essential in order to understand the geographical situation of an area. This paper introduces a new distributed approach to identify as well as locate deforestation across different clients using Federated Learning (FL). Federated Learning enables distributed network clients to collaboratively train a model while maintaining data privacy and security of the active users. In our framework, a client corresponds to an edge satellite center responsible for local data processing. Moreover, FL provides an advantage over centralized training method which requires combining data, thereby compromising with data security of the clients. Our framework leverages the \textbf{FLOWER} framework with \textbf{RAY} framework to execute the distributed learning workload. Furthermore, efficient client spawning is ensured by \textbf{RAY} as it can select definite amount of users to create an emulation environment. Our FL framework uses YOLOS-small (a Vision Transformer variant), Faster R-CNN with a ResNet50 backbone, and Faster R-CNN with a MobileNetV3 backbone models trained and tested on publicly available datasets. Our approach provides us a different view for image segmentation-based tasks on satellite imagery.  
\end{abstract}
\section{INTRODUCTION}
\IEEEPARstart{F}{EDERATED}  learning (FL) is a decentralized machine learning method where multiple clients collaboratively train a model without sharing their local data, ensuring data privacy and reducing communication overhead [\citenum{mcmahan2023communicationefficientlearningdeepnetworks}]. This approach is particularly valuable in scenarios where data cannot be centralized due to privacy concerns, legal restrictions, or bandwidth limitations [\citenum{zhu2017deep}]. FL allows each client to train a model on its own data and send only model updates to a central server, which aggregates these updates to improve a global model, making it suitable for distributed and sensitive applications [\citenum{kairouz2021advances}]. In the context of remote sensing, FL addresses challenges such as data sovereignty and the large volume of satellite imagery, which often contains sensitive information and is impractical to transfer centrally [\citenum{moreno2024federated}]. Among remote sensing applications, detecting deforestation from satellite imagery is critical for understanding and mitigating the impacts of forest loss on biodiversity and climate change, as automated approaches are essential for efficient and accurate monitoring over large areas [\citenum{md2024deforestation}]. Traditional centralized machine learning approaches face significant hurdles, including privacy risks from sharing raw satellite data across organizations and the high communication overhead of transferring large datasets [\citenum{zhang2023local}]. Our paper proposes a federated learning framework for deforestation detection, leveraging the \textbf{FLOWER} \textsuperscript{\ref{foot:flower}} and \textbf{RAY} \textsuperscript{\ref{foot:ray}} frameworks to implement a scalable system, evaluating models like YOLOS-small and Faster R-CNN with different backbones, and using aggregation strategies such as FedAvg and FedAvgM [\citenum{chi2016big}, \citenum{beutel2020FLOWER}]. This approach aims to achieve competitive performance while preserving data privacy and reducing communication costs, as demonstrated through extensive experiments [\citenum{moritz2018ray}]. 

The contributions of this work are summarized as follows:

\begin{itemize}
    \item Development of a novel FL framework for privacy-preserving deforestation detection using satellite imagery, addressing data sovereignty and communication challenges.
    \item Evaluation of three distinct model architectures (YOLOS-small, Faster R-CNN with ResNet50, and Faster R-CNN with MobileNetV3) under two aggregation strategies (FedAvg and FedAvgM) to identify optimal configurations for FL-based deforestation detection.
    \item Comprehensive comparison of FL performance against centralized training, demonstrating competitive accuracy while preserving privacy, particularly at smaller scales.
    \item Analysis of model stability, scalability, and training time trade-offs, providing practical insights for deploying FL in real-world Earth observation tasks .
\end{itemize}

\subsection{Related Works}
The accurate detection and monitoring of deforestation are critical for environmental conservation, climate change mitigation, and preserving biodiversity.  Traditionally, this task involves analyzing Remote Sensing (RS) data, often satellite imagery, using AI techniques like Machine Learning (ML) and Deep Learning (DL)  [\citenum{Haq2024}]. Studies have utilized various models, including CNNs and FCNs (U-Net, SegNet), for change detection and classification [\citenum{Haq2024}]. However, these centralized approaches face significant challenges regarding data privacy, communication overhead, and regulatory compliance [\citenum{Jallepalli2021}, \citenum{liu2020fedvision}].

Federated Learning (FL) offers a privacy-preserving alternative by enabling collaborative model training on decentralized data [\citenum{Jallepalli2021}, \citenum{Zhang}]. In FL, clients train models locally, and only updates are aggregated centrally, often using algorithms like Federated Averaging (FedAvg) [ \citenum{Jallepalli2021}, \citenum{liu2020fedvision}, \citenum{Zhang}].

FL's feasibility for complex computer vision tasks, such as object detection, has been demonstrated. Jallepalli et al. [\citenum{Jallepalli2021}] applied FL with YOLOv3 in autonomous driving, matching centralized performance while ensuring privacy. Liu et al. [\citenum{liu2020fedvision}] deployed FedVision, an FL platform using FedYOLOv3 for commercial visual detection tasks, highlighting real-world benefits like efficiency and cost reduction, alongside techniques like model compression [\citenum{liu2020fedvision}].

Research has also addressed FL's robustness and security. To counter single-point-of-failure and poisoning attacks, Zhang et al. [\citenum{Zhang}] proposed FedVisionBC, integrating blockchain with FL and YOLOv5. Their work also introduced ADPFedAvg, which adds differential privacy to FedAvg to prevent membership inference attacks [\citenum{Zhang}]. Handling statistical heterogeneity (non-IID data) across clients, a common real-world FL challenge [\citenum{Jallepalli2021}, \citenum{liu2020fedvision}], is another focus, with strategies like FedAvgM being developed.

While prior work covers RS/AI for deforestation [\citenum{Haq2024}] and FL for general vision tasks [\citenum{Jallepalli2021}, \citenum{liu2020fedvision}, \citenum{Zhang}], our paper specifically applies \textit{Federated Deep Learning to Deforestation Detection using Satellite Imagery}. We build on these foundations by evaluating specific models (ResNet, Transformer, MobileNet) and aggregation strategies (FedAvg, FedAvgM) within an FL framework (\textbf{FLOWER} and \textbf{RAY}), directly addressing the deforestation challenge in a distributed, privacy-conscious manner.

\subsection{Motivation} 
Timely detection of deforestation is essential for environmental monitoring and climate change mitigation, often relying on machine learning models trained on large satellite imagery datasets [\citenum{Haq2024}]. However, centralized training poses challenges, including privacy concerns and data sovereignty issues, as organizations may be restricted from sharing sensitive imagery [\citenum{moreno2024federated}]. Additionally, transferring vast datasets incurs significant communication and storage costs [\citenum{kairouz2021advances}]. Federated learning addresses these limitations by enabling collaborative training between distributed clients, such as regional data centers, without sharing raw imagery [\citenum{mcmahan2023communicationefficientlearningdeepnetworks}]. Local models are trained on client datasets, and only model parameters are aggregated centrally, ensuring privacy and reducing communication overhead [\citenum{mcmahan2023communicationefficientlearningdeepnetworks}]. This work aims to demonstrate FL’s effectiveness as a scalable, privacy-preserving approach for training deep learning models for deforestation detection using satellite imagery.

\section{Methodology}

 This section presents the methodology for our federated learning (FL) framework designed to detect deforestation using satellite imagery while preserving data privacy. We describe the dataset, preprocessing steps, client-based data partitioning, and the FL workflow, which leverages the \textbf{FLOWER} [\citenum{beutel2020FLOWER}] frameworks to train three deep learning models: YOLOS-small (a Vision Transformer variant), Faster R-CNN with ResNet50 backbone, and Faster R-CNN with MobileNetV3 backbone. Two aggregation strategies—Federated Averaging (FedAvg) [\citenum{Jallepalli2021}] and Federated Averaging with Momentum (FedAvgM) [\citenum{beutel2020FLOWER}] are used to combine local model updates. The methodology is evaluated using Intersection over Union (IoU) as the primary metric. 

\footnotetext[3]{\label{foot:roboflow}https://universe.roboflow.com/}

\subsection{Dataset and Preprocessing}

We have collected satellite imagery dataset from public available source Roboflow Universe \textsuperscript{\ref{foot:roboflow}}. This satellite imagery dataset was of different sizes and had extra annotations and classes that we do not need for our experiment. To tailor the dataset for deforestation detection, a series of preprocessing steps were applied to refine and standardize the data.  

Firstly, since our objective is to identify deforested regions, we extracted only the images labeled as deforestation  or equivalent classes which represents the same. We discarded the remaining unnecessary classes.

Secondly, all the images from these datasets were resized to a uniform dimension to ensure consistency across different models and clients.

Thirdly, federated learning experimentation is data hungry. To conduct the simulation over several number of clients we need sufficient data for each client. For which we did data augmentation, where we used rotation within the range of -15° to +15°. This transformation helped introduce variability in the dataset while maintaining the spatial characteristics of deforestation patterns.

After augmentation and segmentation, the dataset size was expanded to 12,000 images.

\subsection{Client Data Partitioning}

For federated learning, the preprocessed dataset was partitioned among multiple clients, simulating a distributed learning environment. Each client was assigned a subset of images, ensuring a balanced yet non-identically distributed (non-IID) data scenario. This approach closely mimics real-world conditions where different regions may have varied deforestation patterns.

By applying these preprocessing techniques and partitioning strategies, we created a robust dataset suitable for federated deep learning, enabling accurate and privacy-preserving deforestation detection.

\begin{figure}[h!]
    \centering
    \includegraphics[width=0.95\linewidth]{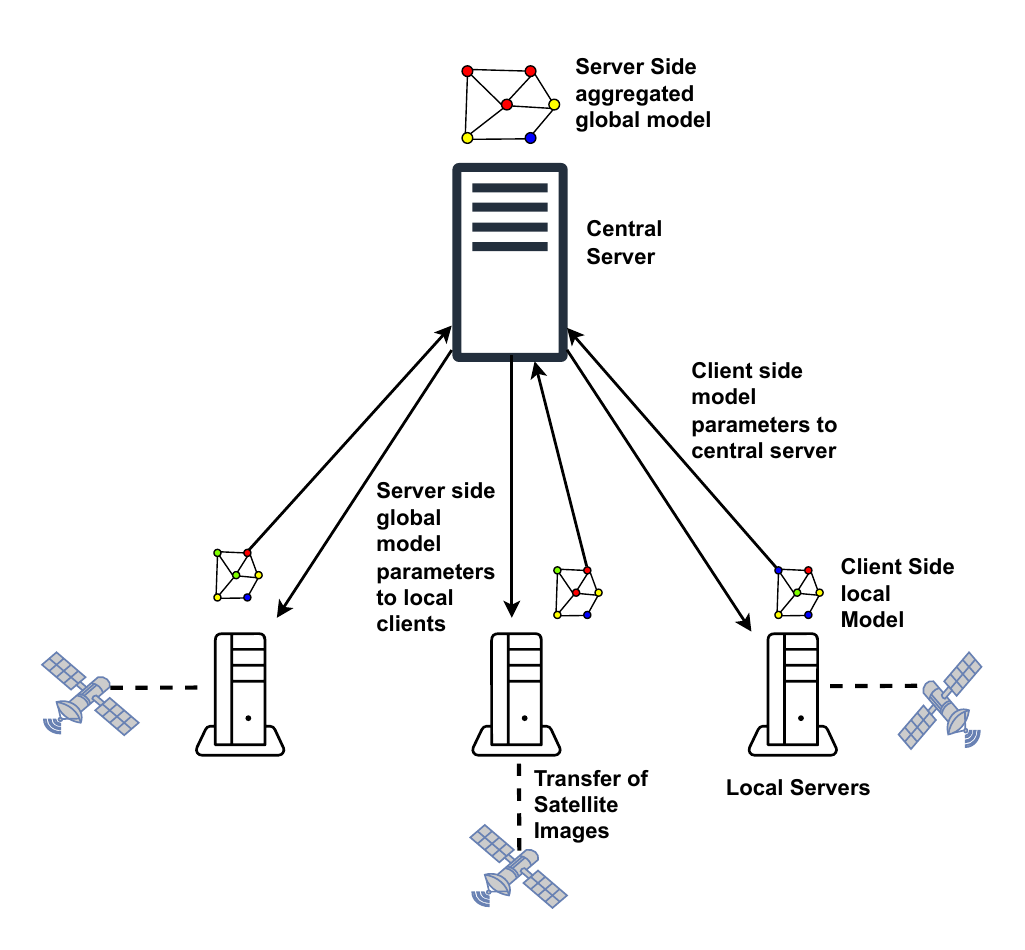}
    \caption{Federated Learning Work Setup}
    \label{FIG:1}
\end{figure}

\begin{figure*}[t] 
    \centering
    \includegraphics[scale=.50]{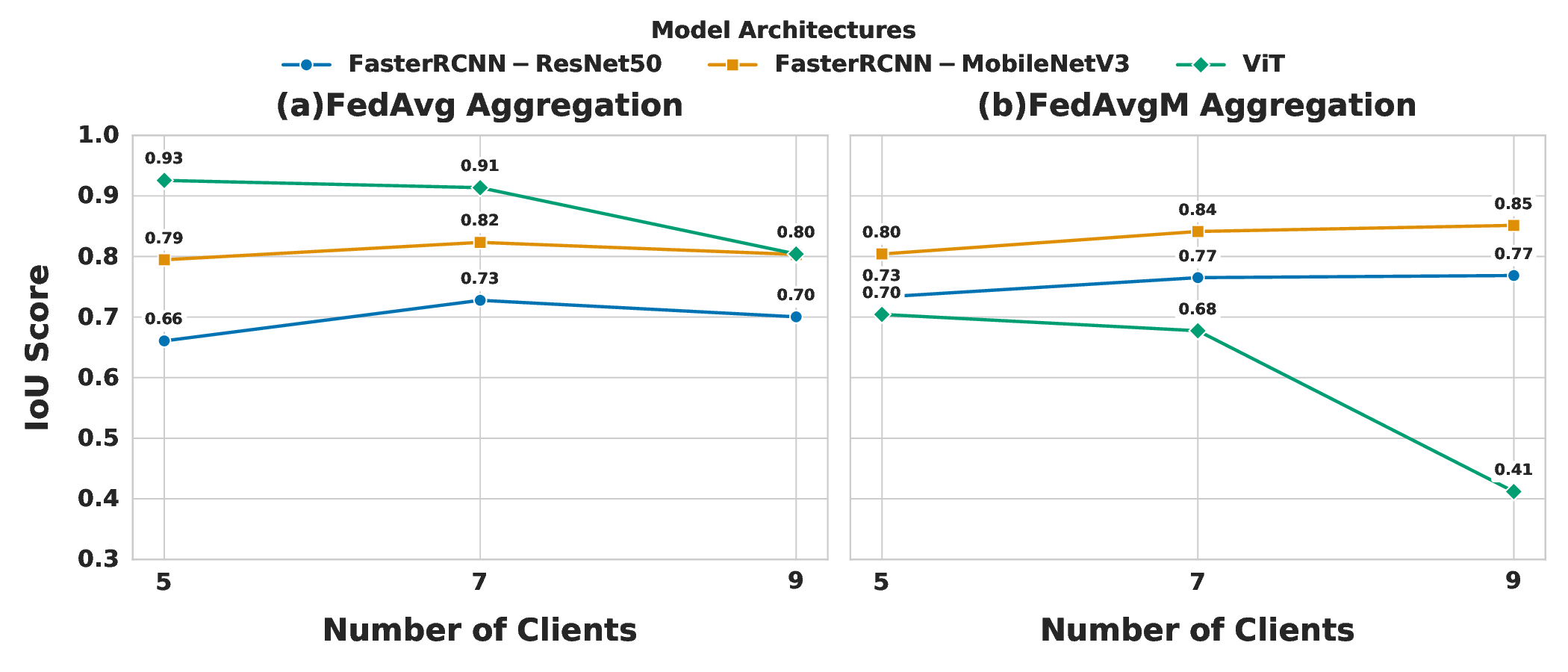} 
    \caption{IOU plots compared with number of clients, models and methods used for aggregation}
    \label{FIG:2}
\end{figure*}

\subsection{Federated Learning}

 The federated learning framework is built on the \textbf{FLOWER} framework, which implements distributed learning simulation across the client and a central service, as shown in Figure \ref{FIG:1}.

 The preprocessed dataset is first grouped into cohorts based on the number of clients. Each cohort is assigned to a different client, simulating a scenario where data is geographically distributed among multiple locations or devices. During each training round a random subset of clients is activated. Here, the \textbf{RAY} framework optimizes resource allocation to balance computational demands and communication efficiency. Each client is allocated 1.00 CPU and 1.00 GPU, corresponding to 2 Intel Xeon vCPUs and 2 NVIDIA Tesla T4 GPUs, through explicit resource specification during deployment.

 Here,  we took three combinations of number of clients, i.e 5, 7, and 9 where 7 global rounds were given to train each. We used 3 models to carry out our experiment:

 \begin{enumerate}
     \item \textbf{\textbf{YOLOS-small} }: Based on the "You Only Look at One Sequence" approach [\citenum{Fang2021}], this model is a variant of the Vision Transformer (ViT) architecture specifically adapted for object detection. We utilize the 'small' version, available via Hugging Face and fine-tuned on COCO 2017, chosen for its relative computational efficiency compared to larger transformer models, making it potentially suitable for clients with varying resource constraints in a federated setting. It processes images as sequences and uses a bipartite matching loss similar to that of DETR. 
     \item \textbf{\textbf{Faster R-CNN with ResNet50 Backbone} }: This represents a widely-used and robust two-stage object detection framework. It employs a ResNet50 [\citenum{He_2016_CVPR}] backbone, a deep convolutional neural network known for strong feature extraction capabilities. We initialize the model using standard pre-trained weights, leveraging features learned on large-scale datasets to potentially enhance performance and speed up convergence during local training on the deforestation task. 
     \item \textbf{\textbf{Faster R-CNN with MobileNetV3 Backbone} }: To assess a more lightweight option within the two-stage framework, we also include Faster R-CNN configured with a MobileNetV3 (Large) [\citenum{Howard_2019_ICCV}] backbone. MobileNetV3 is designed specifically for efficiency, particularly on edge or mobile devices, making this configuration highly relevant for FL scenarios. We utilize the corresponding pre-trained weights to benefit from existing feature representations while evaluating its performance in the distributed setting. 
 \end{enumerate}

Before using federated learning, we trained each of these models using a sample dataset and performed parameter tuning to determine the required learning rate and batch size. This allowed us to achieve optimal training and efficient results. 

Following local training, the model parameters will be transmitted to a central server, where an aggregation method will be used to merge the models from each of those clients. Here, we tried out two different aggregation techniques: 

 \begin{enumerate}
     \item \textbf{Federated Averaging (FedAvg)}:  
FedAvg [\citenum{mcmahan2023communicationefficientlearningdeepnetworks}, \citenum{sun2021decentralizedfederatedaveraging}] is one of the simplest and most widely used strategies in federated learning.  FedAvg calculates a weighted average of local updates, proportional to each client’s sample size. It can be expressed by :

\[
w_{\text{glob}}^{t+1} \leftarrow w_{\text{glob}}^t - \sum_{k \in S_t} \frac{n_k}{n} \Delta w_k^{t+1}
\]

\[
\Delta w_k^{t+1} = w_k^{t+1} - w_{\text{glob}}^t
\]

Here the global model \( w_{\text{glob}}^{t+1} \) is updated as a weighted average of the local model updates \( w_k^{t+1} \), where the weights \( \frac{n_k}{n} \) represent the proportion of data samples in each client relative to the total number of samples in all participating clients.

    \item \textbf{Federated Averaging with Momentum (FedAvgM)}:
FedAvgM [\citenum{hsu2019measuringeffectsnonidenticaldata}] is an enhanced version of the standard FedAvg algorithm that adds momentum on the server side. Like in traditional gradient descent, momentum helps speed up and stabilize convergence. The server maintains a velocity vector that accumulates past updates. Each global model update combines the weighted average of current client updates (as in FedAvg) with this velocity, guiding updates more consistently and reducing oscillations. This leads to faster and smoother convergence, especially in complex optimization scenarios. 
\[
w_{\text{glob}}^{t+1} \leftarrow w_{\text{glob}}^t - \eta v^{t+1}
\]
\[
v^{t+1} = \beta v^t + (1 - \beta) \Delta w^{t+1}
\]

Where:  
\( v^t \) = Server momentum in round \( t \), 
\( \beta \) = Momentum parameter (\( 0 \le \beta < 1 \)), \( \eta \) = Learning rate, \( \Delta w^{t+1} \) = Weighted average of client model updates.
 \end{enumerate}

The aggregated global model is broadcast to all clients, initiating the next training round [\citenum{mcmahan2023communicationefficientlearningdeepnetworks}]. This iterative process persists for a predetermined number of rounds or until convergence is observed, based on the validation performance [\citenum{kairouz2021advances}]. The global model is evaluated on a held-out test set using IoU, which quantifies the overlap between predicted and ground-truth bounding boxes for deforested regions.

\[
\mathrm{IoU} = \frac{\mathrm{Area\ of\ Overlap}}{\mathrm{Area\ of\ Union}}
\]

\begin{figure}[htbp]
    \centering 

    \begin{subfigure}{\columnwidth} 
        \centering 
        \includegraphics[width=0.85\linewidth]{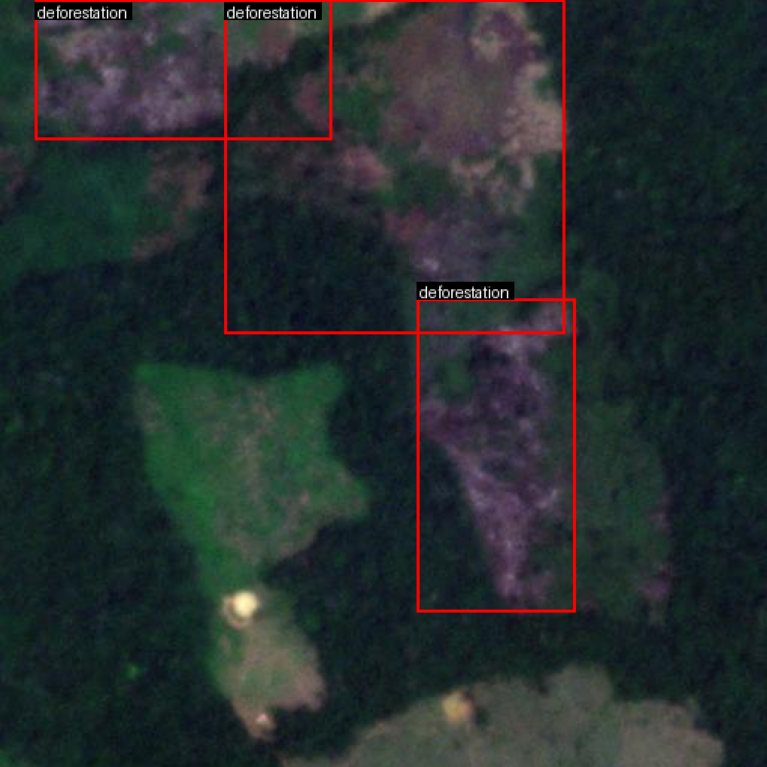} 
        \caption{Original deforestation annotations (Ground Truth).} 
        \label{fig:original_annotations} 
    \end{subfigure}

    \vspace{\medskipamount} 

    \begin{subfigure}{\columnwidth} 
        \centering
        \includegraphics[width=0.9\linewidth]{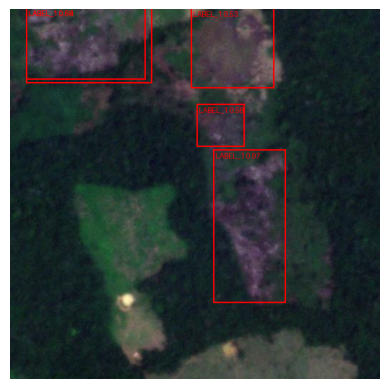} 
        \caption{Model predictions for deforestation.} 
        \label{fig:model_predictions} 
    \end{subfigure}

    \caption{Comparison of deforestation detection. (a) shows the ground truth bounding boxes, and (b) shows the bounding boxes predicted by the model on the same satellite image.}
    \label{fig:detection_comparison} 
\end{figure}
We can compare the ground truth in Figure~\ref{fig:original_annotations} with the model's output shown in Figure~\ref{fig:model_predictions}. The overall comparison is presented in Figure~\ref{fig:detection_comparison}.

\section{RESULTS \& DISCUSSION}
This section presents the experimental results from applying our proposed federated learning framework to the deforestation detection task using satellite imagery. We evaluated the performance, measured by Intersection over Union (IoU), of three distinct model architectures (YOLOS-small, Faster R-CNN with ResNet50, and Faster R-CNN with MobileNetV3) under two different aggregation strategies (FedAvg and FedAvgM). The analysis focuses on the impact of these choices, client scalability, and provides a comparison against traditional centralized training baselines.

\subsection{Performance Analysis}
Our experimental evaluation compared three model architectures and two aggregation strategies using Intersection over Union (IoU), with results shown in Figure \ref{FIG:2}. We selected client counts of 5, 7, and 9 to create a clear progression for evaluating model scalability. Faster R-CNN with a MobileNetV3 backbone consistently performed well, achieving the highest score ($\approx 85.0\%$ IoU) with 9 clients using FedAvgM (Figure \ref{FIG:2}b), indicating its suitability for scaled federated learning due to efficiency and robustness.

Comparing aggregation methods, FedAvgM generally improved performance for the CNN-based models (Faster R-CNN with ResNet50/MobileNetV3) by up to 11\% compared to FedAvg. Conversely, the Vision Transformer (ViT), which attained the overall peak IoU ($\approx 93.0\%$) with FedAvg and 5 clients (Figure 3a), experienced a significant performance drop ($\approx 25\%$) when switched to FedAvgM under the same conditions. This suggests potential interference between momentum aggregation and transformer attention mechanisms in FL.

Regarding scalability, the CNN-based models maintained relatively stable IoU scores as client numbers increased from 5 to 9. In contrast, ViT demonstrated high sensitivity, particularly under FedAvgM, where its performance degraded substantially from 70.0\% (5 clients) to 41.0\% (9 clients). This implies that while transformers can achieve high performance in specific FL scenarios (low client count, FedAvg), CNN architectures may offer better robustness for larger-scale federated deployments based on these results.

\begin{figure}[h!]
    \centering
    \includegraphics[scale=0.28]{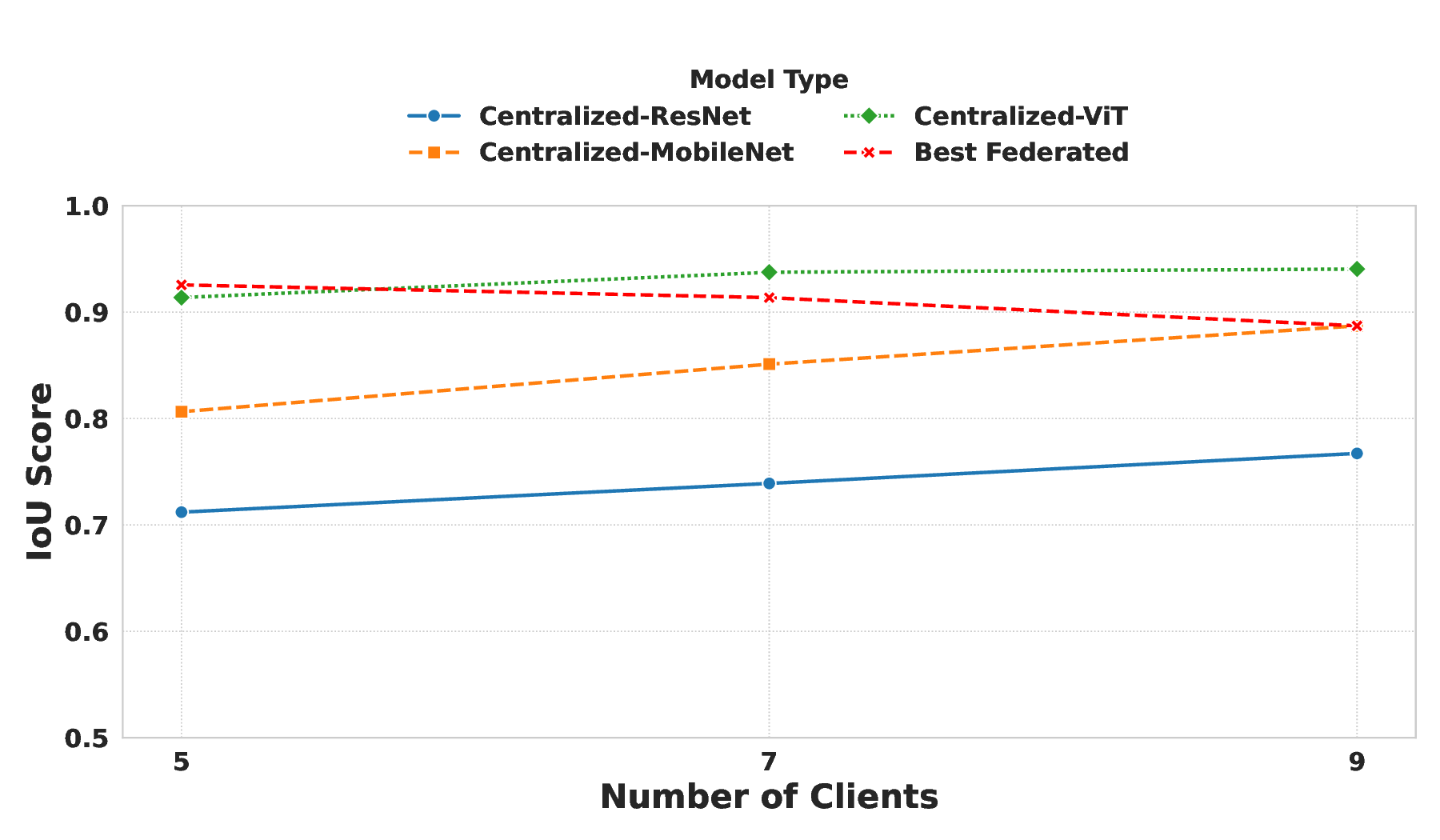}
    \caption{Centralized vs Federated Learning}
\end{figure}

\subsection{Comparison on Baseline (Centralized)}
To benchmark federated performance, we compared the peak FL results (highest IoU achieved by any FL setup per client count) against models trained centrally on combined client data (Figure 4). Centralized training showed performance gains with more aggregated data, with ViT consistently achieving the highest IoU (up to $\approx 94.1\%$), followed by MobileNetV3 (up to $\approx 88.7\%$), and ResNet50 (up to $\approx 76.7\%$) when using data from 9 clients.

Crucially, at the 5-client scale, the best FL configuration (ViT with FedAvg, $\approx 92.6\%$) nearly matched the top centralized performance. As the client count increased, a gap emerged relative to the optimal centralized ViT. However, even at 9 clients, the peak FL result ($\approx 88.7\%$, achieved by MobileNetV3 with FedAvgM) effectively equaled the performance of the centralized MobileNetV3\textit{ model}.

These findings indicate that while centralized training often represents the performance ceiling, well-configured FL can yield highly competitive results, especially at smaller scales, and can match strong centralized alternatives even as the system scales. This highlights the importance of optimizing the FL model and aggregation strategy.

\begin{figure}[h!]
    \centering
    \includegraphics[scale=0.315]{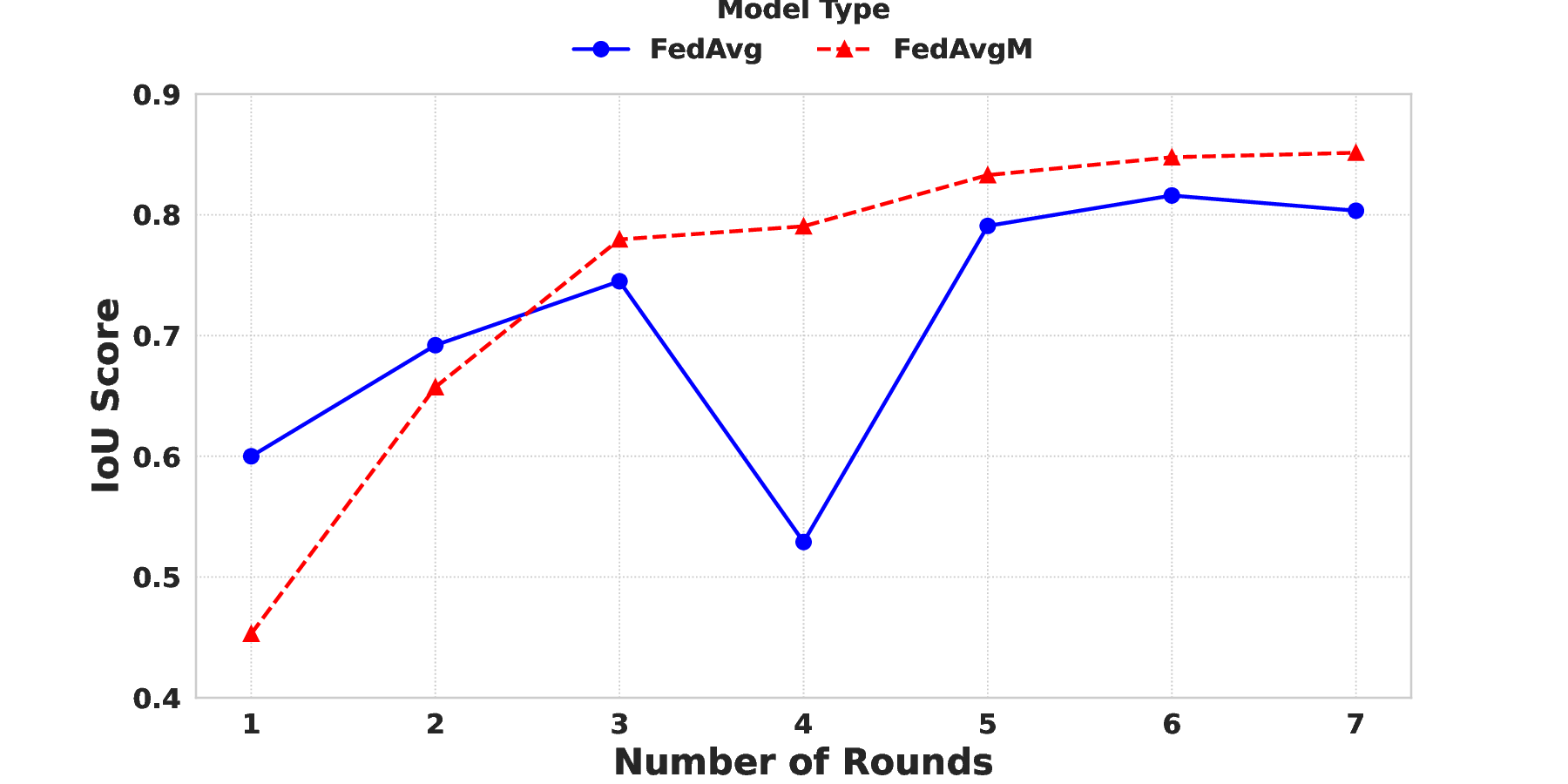}
    \caption{IoU Convergence: FasterRCNN-MobileNetV3 (9 Clients)}
\end{figure}

\subsection{Convergence Analysis}
To further examine the learning dynamics of the best-performing practical architecture identified (FasterRCNN-MobileNetV3) at the largest tested scale (9 clients), we plotted its IoU score progression over the initial 7 communication rounds for both FedAvg and FedAvgM aggregation strategies. Figure 5 illustrates this comparison.

Both strategies demonstrate learning occurs, as indicated by the general upward trend in IoU scores. However, the convergence paths differ significantly. The standard FedAvg strategy (blue solid line) exhibits notable instability; after reaching a peak of approximately 0.745 IoU at round 3, it suffers a sharp drop to 0.529 IoU in round 4 before recovering to around 0.80-0.82 IoU in the later rounds shown.
In contrast, the FedAvgM strategy (red dashed line) displays a much more stable and monotonic improvement. Although starting at a lower IoU score (0.453) in round 1 compared to FedAvg (0.6), FedAvgM consistently increases its performance, surpassing FedAvg after round 3 and reaching the highest IoU score of approximately 0.851 by round 7.

This direct comparison highlights the stabilizing effect of momentum in the FedAvgM algorithm for the FasterRCNN-MobileNetV3 model under these conditions. FedAvgM not only leads to a better final IoU score but also avoids the significant performance instability observed with standard FedAvg. This suggests that FedAvgM contributes to a more reliable and efficient learning process for this CNN-based architecture within our scaled federated setting.

\begin{figure}[h!]
    \centering
    \includegraphics[scale=0.35]{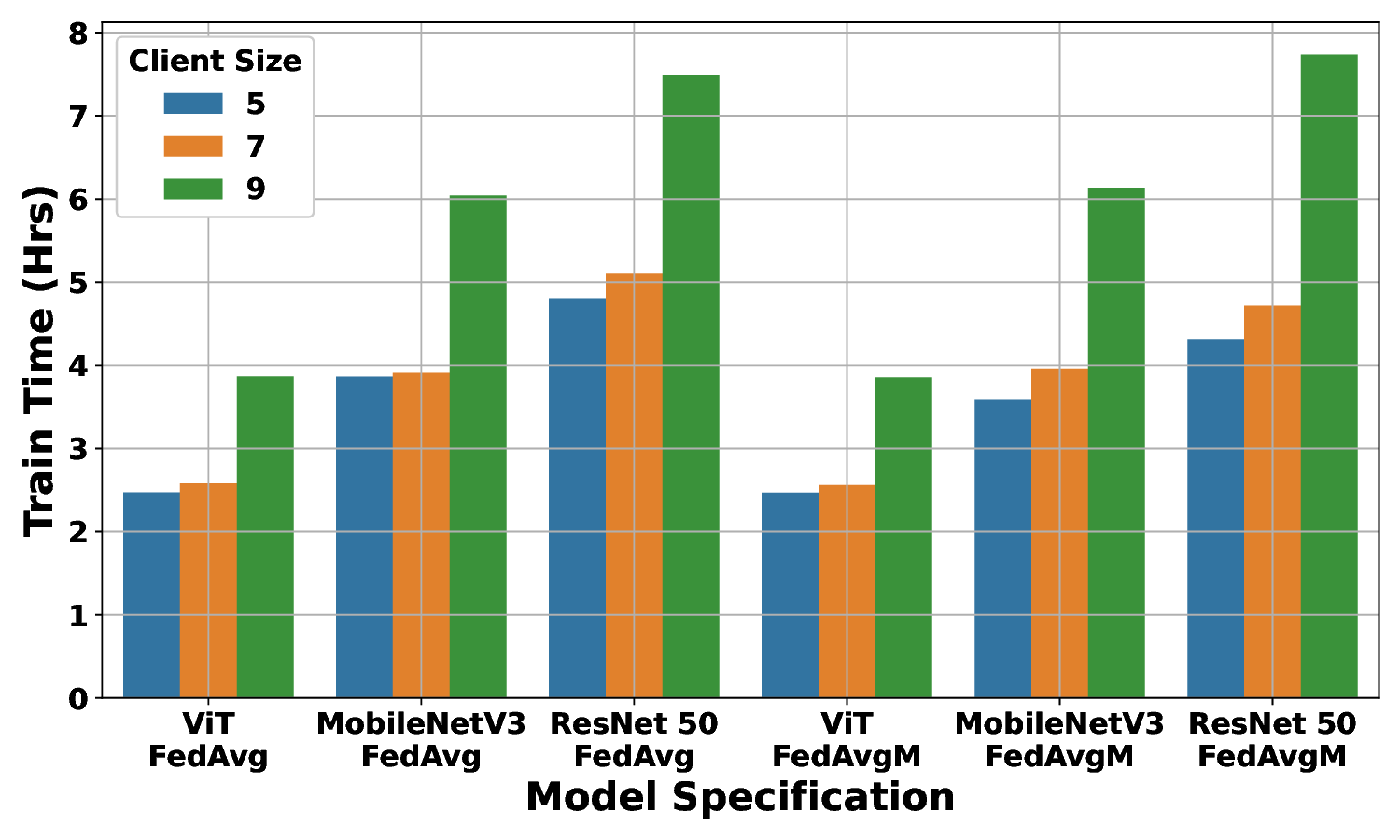}
    \caption{Training Time for different Model Specifications}
\end{figure}

\subsection{Training Time Analysis}
We also evaluated the total wall-clock training time for each configuration (Figure 6). As expected, training time increased with the number of clients for all setups. Architecturally, Faster R-CNN with ResNet50 was consistently the slowest (up to ~7.5 hrs), while the ViT (YOLOS-small) model was the fastest (under 4 hrs). Faster R-CNN with MobileNetV3 required intermediate time (up to ~6.0 hrs). The choice between FedAvg and FedAvgM had a negligible effect on total training time for a given model and client count. Considering both time and accuracy, MobileNetV3 presents a strong practical trade-off, offering robust performance (Section III.A) with moderate computational cost compared to the faster but less stable ViT and the slower ResNet50.

\section{CONCLUSION}

This paper successfully demonstrated the application of Federated Learning (FL) for privacy-preserving deforestation detection using satellite imagery, evaluating YOLOS-small (ViT), Faster R-CNN-ResNet50, and Faster R-CNN-MobileNetV3 with FedAvg and FedAvgM aggregation strategies using the \textbf{FLOWER} and \textbf{RAY} frameworks.

Our results confirm FL's effectiveness for this task. Faster R-CNN with MobileNetV3 emerged as a particularly robust and practical choice, achieving strong and stable performance ($\approx85.0\%$ IOU) across different client counts, especially benefiting from FedAvgM aggregation. Although the ViT model achieved the highest peak IoU ($\approx 93.0\%$) in a low-client FedAvg scenario, it exhibited significant sensitivity to scaling and the FedAvgM strategy, alongside faster training times. This highlights a clear trade-off: ViT offers potential peak performance and speed but lacks the robustness of MobileNetV3, which provided a better balance of high accuracy, stability (particularly with FedAvgM), and moderate training time, making it more suitable for reliable deployment.

The comparison with centralized methods showed FL can achieve highly competitive results, nearly matching the best centralized performance at smaller scales and equaling strong centralized alternatives (like MobileNetV3) at larger scales, provided the FL configuration is chosen carefully. Overall, this work confirms FL's viability for sensitive Earth observation tasks, underscoring the critical impact of model architecture and aggregation strategy choices on performance, scalability, and efficiency. MobileNetV3 offers a promising pathway for practical deployment, while further research is needed to optimize transformer architectures for robust, large-scale federated learning.

\bibliographystyle{IEEEtran}

\bibliography{Ref}

\end{document}